\PassOptionsToPackage{table}{xcolor}
\documentclass[letterpaper,10pt,conference]{ieeeconf}

\IEEEoverridecommandlockouts
\usepackage[hyphens]{url}
\usepackage{graphicx}
\usepackage{svg}
\usepackage{xcolor}
\usepackage{cite}
\usepackage{booktabs}
\usepackage{multirow}
\usepackage{amsmath}
\usepackage{amssymb}

\newcommand{\includesvgasset}[3][]{%
  \IfFileExists{#2}{\includegraphics[#1]{#3}}{%
    \PackageError{SimFuse3D}{Missing SVG source #2}{Restore the corresponding SVG figure.}}}
\title{SimFuse3D: Source-Guided Target Simulation and Confidence-Guided Multi-Stage Localization Reweighting for Cross-Platform 3D Object Detection}
\author{Yongchun Lin$^{1}$, Xinliang Zhang$^{2}$, Yun Zou$^{1}$, Zhixuan Xiao$^{1}$, Liang Lei$^{1,*}$, \\
Jianya Guo$^{2}$, Yuqiang Zhai$^{2}$, Xiaofeng Wang$^{2}$, HaiKuo Xu$^{2}$, 
and Haoang Li$^{3,*}$%
\thanks{This work was supported by the Natural Science Foundation of Guangdong Province, China, under Grant 2023A1515010206.}%
\thanks{$^{*}$Corresponding authors: Liang Lei and Haoang Li.}%
\thanks{$^{1}$Yongchun Lin, Yun Zou, Zhixuan Xiao, and Liang Lei are with Guangdong University of Technology.
        {\tt\small \{linyongchun,zouyun1, xiaozhixuan\}@mails.gdut.edu.cn; leiliang@gdut.edu.cn}}%
\thanks{$^{2}$Xinliang Zhang, Jianya Guo, Yuqiang Zhai, Xiaofeng Wang, and Haikuo Xu are with SenseTime.
        {\tt\small \{zhangxinliang, guojianya, zhaiyuqiang, wangxiaofeng, xuhaikuo\}@senseauto.com}}%
\thanks{$^{3}$Haoang Li is with The Hong Kong University of Science and Technology (Guangzhou).
        {\tt\small haoangli@hkust-gz.edu.cn}}%
}

\begin{document}
\maketitle
\thispagestyle{empty}
\pagestyle{empty}

\begin{abstract}
Changes in sensor height and viewpoint alter object-level point distributions, making cross-platform LiDAR unsupervised domain adaptation (UDA) difficult. Self-training uses labeled source scans and unlabeled target scans, yet a retained prediction may provide a useful target location while enclosing sparse foreground returns, background clutter, or points inconsistent with the predicted box. We refer to this mismatch as \emph{box--point inconsistency}. We introduce \textbf{SimFuse3D}, which preserves the target placement and repairs the associated pseudo object using measured geometry from labeled source scans. \emph{Object Memory} retrieves a similar labeled source instance. \emph{Target Simulation} places the retrieved source geometry at the target location, aligns its points with the target viewing geometry, and filters the aligned crop to approximate the target observation. \emph{Confidence-Guided Multi-Stage Localization Reweighting} (CMLR) maps each target pseudo-object confidence score to a bounded weight shared by RPN localization and R-CNN box regression. All components operate only during adaptation, leaving the detector architecture and inference graph unchanged. Across six cross-platform transfers, SimFuse3D consistently outperforms Pi3DET-Net and achieves the best performance among the compared adaptation methods on nearly all metrics. On nuScenes$\rightarrow$KITTI, it ranks first among the compared adaptation methods with both evaluated detectors.
\end{abstract}
\section{Introduction}
\label{sec:introduction}

LiDAR-based 3D object detection supports autonomous vehicles and mobile robots. Most public benchmarks, including KITTI~\cite{geiger2012kitti} and nuScenes~\cite{caesar2020nuScenes}, are collected from road vehicles. As documented by Pi3DET~\cite{Liang2025Pi3DET}, a detector transferred to a drone or quadruped robot encounters a different sensor height, motion pattern, and target distribution. More importantly for point-based training, viewpoint changes alter the spatial distribution and completeness of object-level observations. Collecting accurate 3D boxes for every platform remains expensive.

\begin{figure}[!t]
  \vspace*{5pt}
  \centering
  \includesvgasset[width=\columnwidth]{figures/teaser_visio_v30_a.svg}{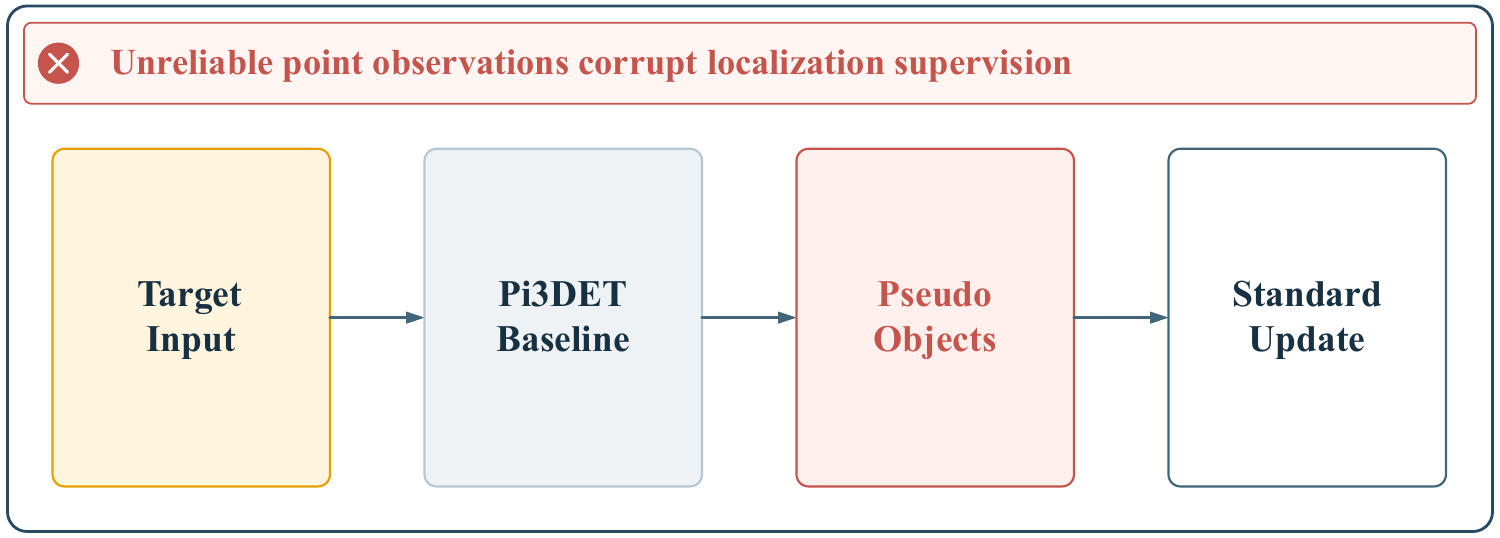}\par
  \vspace{0.35mm}
  {\footnotesize\textbf{(a) Pi3DET-Net baseline}}\par
  \vspace{1.0mm}
  \includesvgasset[width=\columnwidth]{figures/teaser_visio_v30_b.svg}{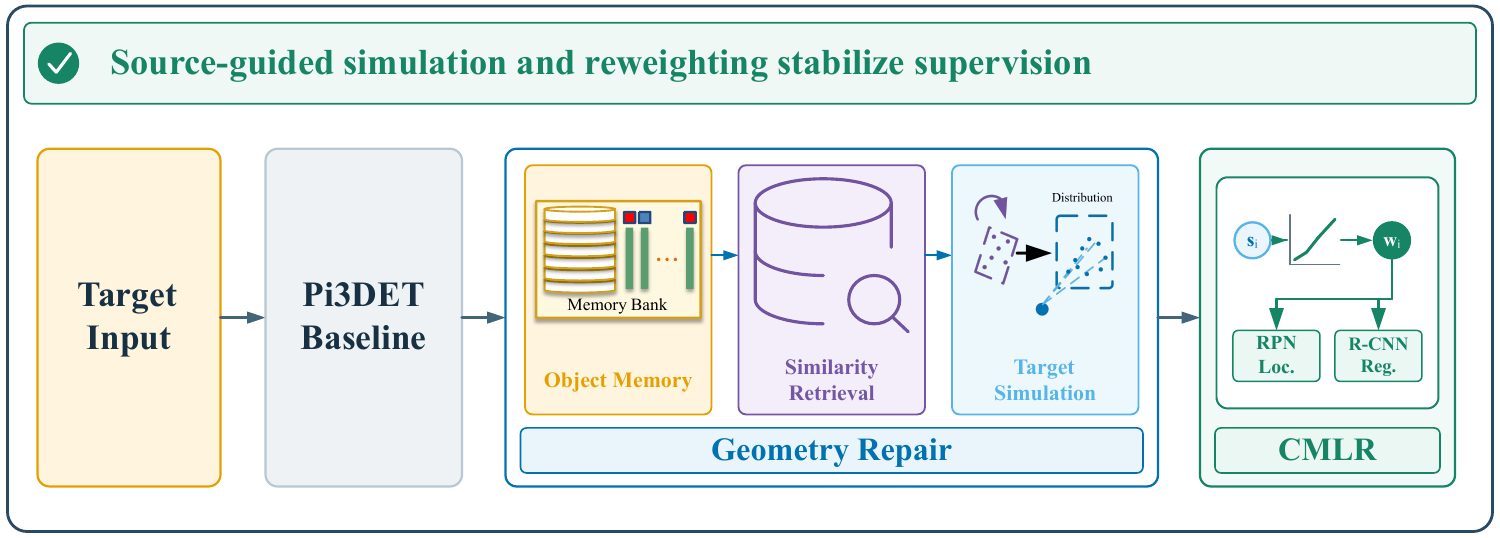}\par
  \vspace{0.35mm}
  {\footnotesize\textbf{(b) SimFuse3D adaptation}}\par
  \vspace{1.0mm}
  \includesvgasset[width=\columnwidth]{figures/drone_fig_c_00153.svg}{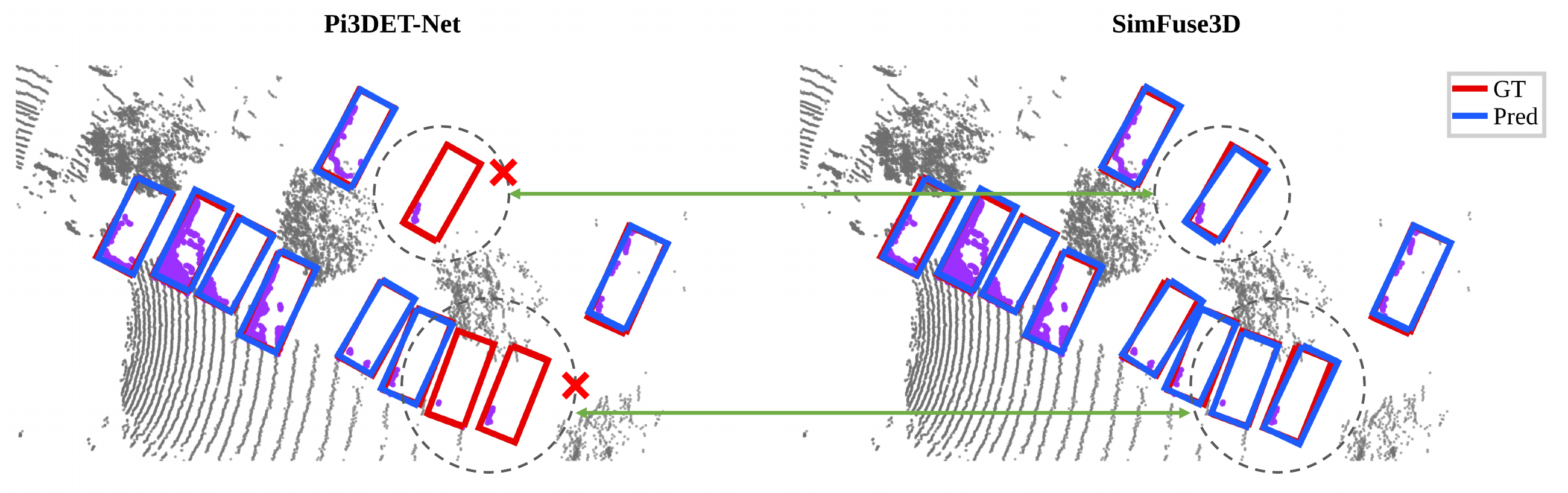}\par
  \vspace{0.35mm}
  {\footnotesize\textbf{(c) Vehicle$\rightarrow$Drone detection}}\par
  \vspace{1.0mm}
  \caption{\textbf{Motivation and qualitative comparison.} \textbf{(a)} Pi3DET-Net~\cite{Liang2025Pi3DET} retains the target point observations enclosed by accepted pseudo boxes, allowing unreliable geometry to enter localization supervision. \textbf{(b)} Our method, SimFuse3D, retrieves measured source geometry, aligns it with the target placement and viewpoint, and reweights target localization supervision with CMLR. \textbf{(c)} Vehicle$\rightarrow$Drone example. The upper dashed green circle marks one ground-truth vehicle, and the lower circle marks two adjacent vehicles. All three boxes contain only a few LiDAR returns. Pi3DET-Net misses the three vehicles in the two marked regions, whereas SimFuse3D detects them. Red boxes show ground truth, blue boxes show predictions, and purple points are returns inside the ground-truth boxes.}
  \label{fig:motivation}
\end{figure}

Unsupervised domain adaptation (UDA) offers a practical alternative to annotating each target platform: it combines labeled source scans with unlabeled target scans. In the cross-platform setting introduced by Pi3DET~\cite{Liang2025Pi3DET}, we use Pi3DET-Net as the self-training baseline. Methods such as ST3D~\cite{Yang2021ST3D}, ST3D++~\cite{Yang2023ST3DPP}, and MS3D++~\cite{Tsai2025MS3DPP} improve target supervision by filtering predicted boxes or refining them over time. Yet the points enclosed by an accepted box are usually retained as observed. Across platforms, this can leave a usable box paired with very few foreground returns, points concentrated on one side, or substantial background clutter. We refer to this mismatch as \emph{box--point inconsistency}. It can corrupt localization and RoI supervision even when the pseudo-box location remains useful, and confidence filtering alone cannot correct the underlying observation. SimFuse3D constructs its retrieval memory directly from the labeled source scans already available in UDA. Figure~\ref{fig:motivation}(a)--(b) contrasts the unchanged-point update in Pi3DET-Net with our source-guided geometry repair and confidence weighting; the corresponding cross-platform results are reported in Tables~\ref{tab:cross_platform_results} and~\ref{tab:bidirectional_platform_results}.

\begin{figure*}[!t]
  \vspace*{5pt}
  \centering
  \includesvgasset[width=.98\textwidth,trim=5pt 0pt 0pt 0pt,clip]{figures/framework.svg}{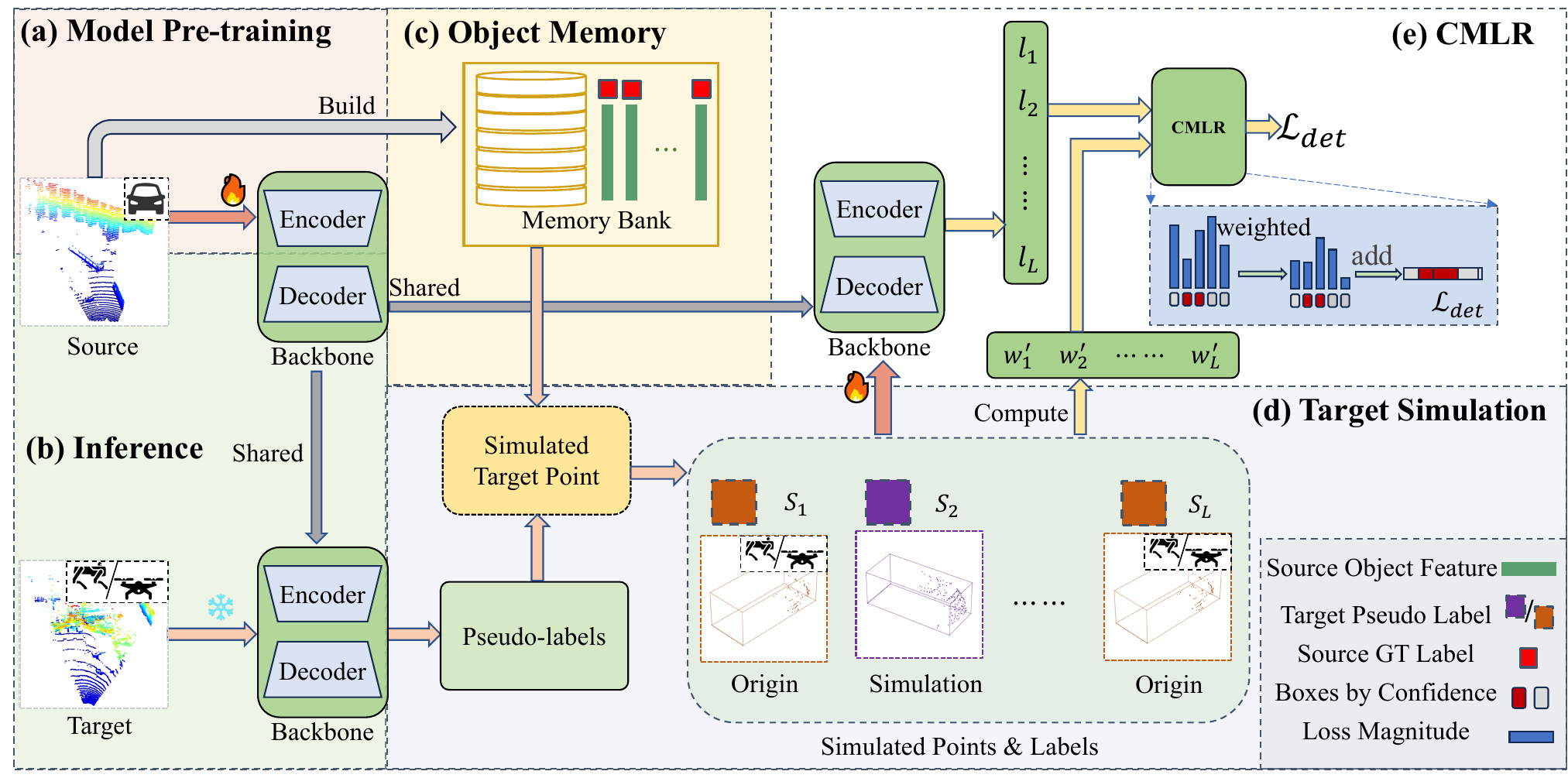}
  \caption{
  \textbf{Overview of our SimFuse3D framework}. (a) The detector is initialized by source-domain pre-training, and (b) target inference generates pseudo labels for adaptation. During adaptation, (c) Object Memory retrieves labeled source instances for target pseudo objects, (d) Target Simulation repairs target pseudo objects with retrieved source geometry, and (e) CMLR reweights localization supervision according to pseudo-object confidence. The modules in (c)–(e) are used only during adaptation and are absent from the deployed inference graph.}
  \label{fig:framework}
\end{figure*}

SimFuse3D separates pseudo-box placement from the enclosed point observation during adaptation. The target prediction supplies the placement, while a retrieved source instance provides measured foreground geometry and object dimensions. Object Memory selects this instance by similarity consensus. Target Simulation aligns it with the target position and viewing direction, filters the aligned crop by target-view angular limits, and replaces the inconsistent in-box observation. Object dimensions come from the retrieved source annotation rather than target-domain size statistics~\cite{Wang2020SN}. If no suitable match is found, the original pseudo object is kept. CMLR maps every target pseudo-object score to a bounded weight shared by RPN localization and R-CNN box regression. Simulated objects receive unit weight, whereas unmatched objects are reweighted from their retained teacher scores. These operations are confined to adaptation; the detector and its inference graph are unchanged (Fig.~\ref{fig:framework}).

This paper makes three contributions:
\begin{itemize}
  \item We identify box--point inconsistency in cross-platform self-training, where a usable pseudo-box location is paired with sparse, contaminated, or geometrically inconsistent points; confidence and point support alone do not fully indicate pseudo-object quality.
  \item We propose SimFuse3D to repair and reweight target pseudo objects. Object Memory and Target Simulation replace unreliable observations with target-view-aligned source geometry, while CMLR applies bounded score-derived weights to all target pseudo objects at the RPN and R-CNN localization stages. The design changes neither the detector architecture nor its inference graph.
  \item Across six transfers and two detectors, SimFuse3D improves all evaluated AP components over Pi3DET-Net and ranks first among the listed adaptation methods in 47 of the 48 comparisons; adding CMLR improves all eight TS-only ablation metrics.
\end{itemize}

\section{Related Work}

\subsection{Unsupervised Domain Adaptation for 3D Detection}
Prior LiDAR adaptation methods reduce domain discrepancy at several levels. Statistical normalization~\cite{Wang2020SN} corrects object-size bias, SPG~\cite{Xu2021SPG} generates semantic foreground points, and GPA-3D~\cite{Li2023GPA3D} aligns object features through geometry-aware prototypes. Sensor changes motivate a different set of techniques: CL3D~\cite{Peng2023CL3D} aligns spatial geometry and temporal motion across configurations, whereas DTS~\cite{Hu2023DTS} combines beam resampling with cross-density consistency. These methods operate on size statistics, representations, or scan density rather than explicitly repairing the box-interior observation of a retained target pseudo object.

Target-supervision methods address a complementary part of the problem. ST3D~\cite{Yang2021ST3D} maintains pseudo labels over time, and ST3D++~\cite{Yang2023ST3DPP} strengthens their denoising and training. ReDB~\cite{Chen2023ReDB} selects pseudo labels according to reliability, diversity, and class balance. MS3D++~\cite{Tsai2025MS3DPP} combines multiple source experts with temporal refinement, although it assumes a multi-source setting. Object-level processing is more closely related to our work: PERE~\cite{Zhang2024PERE} removes points from unreliable boxes or substitutes high-confidence target instances, DALI~\cite{Lu2024DALI} synthesizes points from prepared object-model and sensor libraries, and DiffRefine~\cite{Shin2025DiffRefine} uses diffusion to densify sparse proposals before second-stage refinement. In SimFuse3D, replacement geometry is retrieved directly from the labeled source scans and aligned with the target placement and viewing direction. The target pseudo box determines the insertion location, while the retrieved source instance provides the measured points and ground-truth box dimensions; no separate CAD or sensor-model library is needed.

The information available during adaptation also differs across methods. Attentive Prototypes~\cite{Hegde2024Attentive} assumes that source data are no longer accessible. CMDA~\cite{Chang2024CMDA} uses synchronized images as a semantic bridge, while MMAssist~\cite{Zhao2026MMAssist} combines image and text features and augments pseudo labels with an external 2D detector. These settings are relevant to cross-domain detection, but they are not equivalent to the single-source, LiDAR-only protocol used by SimFuse3D.

\subsection{Point Cloud Representation and Cross-Platform Adaptation}

PointNet~\cite{Qi2017PointNet} and PointNet++~\cite{Qi2017PointNetPP} introduced global and hierarchical point-set representations. Point-NN~\cite{Zhang2023PointNN} later showed that sampling, neighborhood grouping, encoding, and pooling also yield useful non-parametric descriptors. We use this descriptor only to retrieve source instances; it is not part of the detector. Cross-platform evaluation is supported by M3ED~\cite{Chaney2023M3ED}, which contains synchronized vehicle, quadruped, and drone data, and by Pi3DET~\cite{Liang2025Pi3DET}, which provides the corresponding 3D detection benchmarks. UADA3D~\cite{Wozniak2024UADA3D} studies a related setting with sparse LiDAR and large platform gaps. Our focus is narrower: the mismatch between a target pseudo box and the points used to supervise it.

\section{Method}
\label{sec:method}

\subsection{Problem Formulation and Overview}
Let $\mathcal{D}_s=\{(P_s^n,Y_s^n)\}_{n=1}^{N_s}$ denote a labeled source domain, where $P_s^n$ is a point cloud and $Y_s^n$ contains the corresponding 3D ground-truth boxes. Let $\mathcal{D}_t=\{P_t^m\}_{m=1}^{N_t}$ denote an unlabeled target domain collected by a different sensing platform or sampled from a different dataset. Following the self-training protocol of Pi3DET~\cite{Liang2025Pi3DET}, the detector periodically generates pseudo labels for target scan $P_t^m$:
\begin{equation}
 \hat Y_t^m=\{(\hat b_i,\hat y_i,s_i)\}_{i=1}^{N_m},
 \label{eq:pseudo_labels}
\end{equation}
where $N_m$ is the number of retained predictions. Each pseudo label consists of a box $\hat b_i=(x_i,y_i,z_i,l_i,w_i,h_i,\theta_i)\in\mathbb{R}^7$, a class label $\hat y_i$, and a detection score $s_i\in[0,1]$. We use $O_i^t=\{p\in P_t^m\mid p\in\hat b_i\}$ to denote the target points enclosed by $\hat b_i$. In the target domains considered here, some retained boxes contain only a few points; the observed points may cluster on one side of the box, and their outline may not fit the predicted dimensions. We use \emph{box--point inconsistency} to describe these cases. This concern is particularly relevant early in self-training, when target predictions generally have lower scores. A predicted box can therefore provide a useful location cue without providing an equally reliable object observation for training.

At each scheduled pseudo-label refresh, SimFuse3D first queries Object Memory with $O_i^t$. If the retrieval gate accepts a source object, Target Simulation removes the original points inside $\hat b_i$ and inserts the matched source observation after aligning it with the target placement and viewing geometry. This produces a modified scan $\widetilde P_t^m$ and pseudo labels $\widetilde Y_t^m$, combining the target location with measured source geometry from the retrieved instance. If retrieval fails, we retain the original pseudo label. A missing match only indicates that the source memory does not contain a suitable instance, not that the target prediction is incorrect; the original pseudo label may still be correctly localized and provide useful supervision. During the subsequent detector update, CMLR converts pseudo-label scores into bounded object weights for RPN localization and R-CNN box regression. Operations outside these steps follow the Pi3DET-Net baseline.

\begin{figure*}[!t]
  \vspace*{5pt}
  \centering
  \includesvgasset[width=.98\textwidth]{figures/framework2detail.svg}{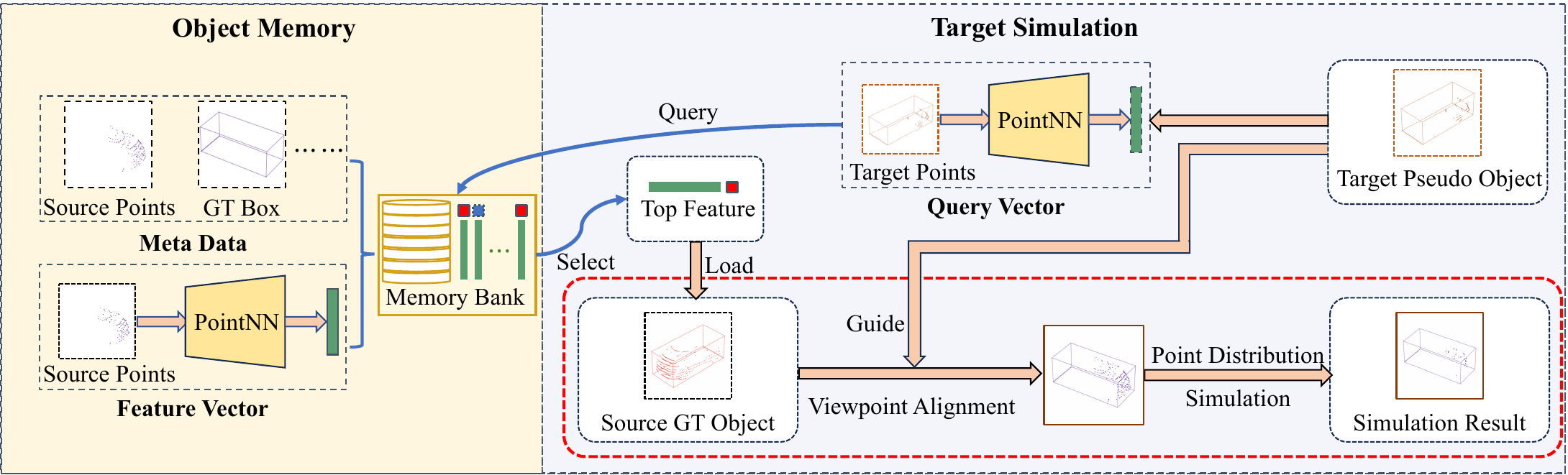}
  \caption{
  Object Memory and the \textbf{proposed Target Simulation}. Object Memory stores source points, ground-truth boxes, metadata, and fixed Point-NN~\cite{Zhang2023PointNN} descriptors for source-instance retrieval. A target pseudo object queries the memory for a labeled source instance. Target Simulation uses the target pseudo object to guide viewpoint alignment and point-distribution simulation, producing the simulated target observation.}
  \label{fig:simulation_detail}
\end{figure*}

\subsection{Object Memory}
\noindent\textbf{Memory construction.}
Object Memory is constructed offline from the labeled source training scans. For each source ground-truth box $b_j^s$, we record its class $y_j^s$ and the corresponding local LiDAR observation $O_j^s$. Both memory entries and target queries use a fixed input of 256 local points. A crop containing more than 256 points is reduced by farthest-point sampling (FPS). For a sparse crop, we enlarge the box by a fixed metric margin to collect nearby measurements before sampling. The xyz coordinates are then translated by the box center but are not rescaled by the box dimensions. We use the fixed Point-NN encoder~\cite{Zhang2023PointNN} only as a non-parametric descriptor $g(\cdot)$ for retrieval:
\begin{equation}
 \begin{aligned}
 z_j^s &= \frac{g(O_j^s,b_j^s)}{\lVert g(O_j^s,b_j^s)\rVert_2},\\
 \mathcal{M} &= \{(z_j^s,b_j^s,O_j^s,y_j^s,m_j)\}_{j=1}^{N_{\mathcal M}}.
 \end{aligned}
 \label{eq:memory}
\end{equation}
Here, $z_j^s$ is a 2,304-dimensional unit vector used only for retrieval. It is computed from raw point coordinates and is not concatenated with detector backbone features. Thus, Object Memory requires no additional retrieval training during adaptation. Each entry also stores the source ground-truth box, point record, class label, frame and box indices, and the source LiDAR origin in $m_j$. Target Simulation uses these records to reload the points inside the source box and recover their sensor-relative viewing direction. The memory contains source instances belonging to the categories used by the target training task. The construction supports multiple categories, although the experiments in this paper use a single car category. The memory is built once and remains fixed throughout target-domain adaptation.

\noindent\textbf{Gated retrieval.}
At each scheduled pseudo-label update, the points associated with a retained target candidate are centered at its predicted box and encoded in the same way to form a unit query $q_i$. In our implementation, an HNSW index produces the initial candidate set, after which the candidates are re-ranked by cosine similarity $a_{ij}=q_i^\top z_j^s$. Let $\pi_i(k)$ denote the index of the $k$-th nearest entry, and let $\mathcal{N}_i$ contain the top $K$ valid entries whose similarity is at least $\delta_n$. Candidate $i$ is associated with a source object only if
\begin{equation}
 \begin{aligned}
 a_{i\pi_i(1)} &\geq \delta_1,\\
 \operatorname{vote}\!\left(\{y_j^s\mid j\in\mathcal{N}_i\}\right) &> 0.
 \end{aligned}
 \label{eq:retrieval_gate}
\end{equation}

The two conditions require a sufficiently similar nearest source instance and a positive majority vote among the qualified neighbors. When both hold, the top-ranked memory entry is passed to Target Simulation and the voted class is assigned to the candidate. Because the inserted object points and replacement box both come from a labeled source instance, we set the replacement confidence to $1.0$. Retrieval is based on Point-NN similarity and neighbor voting; no additional score shield or box-IoU gate is used. If the gate fails, the original positive pseudo label and its teacher score are retained. A failed match means that the current memory does not contain a suitable source instance, rather than that the target pseudo label is necessarily incorrect, so it should not be removed merely because retrieval fails.

\subsection{Target Simulation}
\label{sec:simulation}

\begin{table*}[!t]
\vspace*{6pt}
\caption{Cross-platform adaptation to Pi3DET Quadruped and Drone. Entries are $\mathrm{AP}_{\mathrm{BEV}}/\mathrm{AP}_{\mathrm{3D}}$ (\%). Entries within each detector column share the detector and evaluation configuration; Pi3DET-Net is reproduced under this setup. ST3D/ST3D++ use random object scaling unless $\ddagger$ marks no ROS. ``Target Platform'' is fully supervised; the best scores among the listed adaptation methods are bolded.}
\label{tab:cross_platform_results}
\centering
\small
\setlength{\tabcolsep}{0.6pt}
\renewcommand{\arraystretch}{1.08}
{\spaceskip=1.2pt plus 0.2pt minus 0.1pt\relax
\begin{tabular*}{\textwidth}{@{\extracolsep{\fill}}clcccccccc@{}}
\toprule
\multirow{3}{*}{\shortstack[c]{\textbf{Source}\\\textbf{Domain}}} & \multicolumn{1}{c}{\multirow{3}{*}{\textbf{Method}}} & \multicolumn{4}{c}{\textbf{Pi3DET (Quadruped)}} & \multicolumn{4}{c}{\textbf{Pi3DET (Drone)}} \\
\cmidrule(lr){3-6}\cmidrule(lr){7-10}
& & \multicolumn{2}{c}{\textbf{PV-RCNN}} & \multicolumn{2}{c}{\textbf{Voxel R-CNN}} & \multicolumn{2}{c}{\textbf{PV-RCNN}} & \multicolumn{2}{c}{\textbf{Voxel R-CNN}} \\
\cmidrule(lr){3-4}\cmidrule(lr){5-6}\cmidrule(lr){7-8}\cmidrule(lr){9-10}
& & \textbf{AP@0.7} & \textbf{AP@0.5} & \textbf{AP@0.7} & \textbf{AP@0.5} & \textbf{AP@0.7} & \textbf{AP@0.5} & \textbf{AP@0.7} & \textbf{AP@0.5} \\
\midrule
\multirow{10}{*}{\rotatebox[origin=c]{90}{\textbf{nuScenes}}} & Source Only & \cellcolor{black!6}40.11 / 31.23 & \cellcolor{black!6}43.95 / 42.00 & \cellcolor{black!6}41.84 / 33.39 & \cellcolor{black!6}45.27 / 43.36 & \cellcolor{black!6}35.56 / 24.96 & \cellcolor{black!6}38.36 / 35.76 & \cellcolor{black!6}37.84 / 24.21 & \cellcolor{black!6}44.53 / 39.60 \\
\cmidrule(lr){2-10}
& ST3D~\cite{Yang2021ST3D} & 54.83 / 42.66 & 58.94 / 56.80 & 50.91 / 41.38 & 53.25 / 51.69 & 51.56 / 31.54 & 58.35 / 51.54 & 52.20 / 33.53 & 56.85 / 51.29 \\
& ST3D$^{\ddagger}$~\cite{Yang2021ST3D} & 54.68 / 43.34 & 58.64 / 57.78 & 50.46 / 40.95 & 53.02 / 51.20 & 50.94 / 31.84 & 57.68 / 51.07 & 51.95 / 33.25 & 56.65 / 51.14 \\
& ST3D++~\cite{Yang2023ST3DPP} & 56.08 / 41.90 & 58.84 / 58.03 & 51.05 / 41.45 & 53.76 / 51.70 & 51.26 / 33.35 & 56.63 / 51.34 & 54.32 / 34.70 & 59.41 / 54.03 \\
& ST3D++$^{\ddagger}$~\cite{Yang2023ST3DPP} & 52.48 / 40.41 & 56.56 / 54.17 & 50.90 / 41.42 & 53.12 / 50.97 & 53.03 / 34.18 & 58.45 / 53.02 & 52.97 / 34.79 & 58.54 / 52.52 \\
& ReDB~\cite{Chen2023ReDB} & 52.82 / 41.80 & 56.16 / 54.44 & 51.67 / 40.21 & 54.72 / 52.84 & 52.34 / 34.24 & 57.95 / 52.65 & 50.85 / 34.69 & 54.63 / 49.87 \\
& MS3D++~\cite{Tsai2025MS3DPP} & 57.36 / 45.28 & 61.02 / 59.08 & 52.94 / 43.79 & 55.68 / 55.16 & 55.59 / 33.78 & 61.20 / 55.57 & 55.02 / 36.41 & 59.95 / 54.35 \\
& Pi3DET-Net~\cite{Liang2025Pi3DET} & 56.60 / 45.57 & 61.37 / 59.45 & 54.31 / 43.99 & 57.15 / 55.25 & 55.30 / 37.37 & 62.30 / 55.13 & 55.72 / 35.95 & 60.22 / 55.15 \\
& \textbf{Ours} & \textbf{58.19} / \textbf{47.15} & \textbf{63.02} / \textbf{61.08} & \textbf{57.31} / \textbf{46.49} & \textbf{61.43} / \textbf{59.63} & \textbf{57.32} / \textbf{37.81} & \textbf{62.82} / \textbf{56.82} & \textbf{56.68} / \textbf{38.54} & \textbf{62.18} / \textbf{56.24} \\
\cmidrule(lr){2-10}
& Target Platform & 54.15 / 40.24 & 67.67 / 46.11 & 58.63 / 54.96 & 70.04 / 66.14 & 54.90 / 39.74 & 68.52 / 46.53 & 56.46 / 55.19 & 70.67 / 61.42 \\
\midrule
\multirow{10}{*}{\rotatebox[origin=c]{90}{\textbf{Pi3DET (Vehicle)}}} & Source Only & \cellcolor{black!6}47.57 / 36.73 & \cellcolor{black!6}50.99 / 49.04 & \cellcolor{black!6}50.34 / 37.74 & \cellcolor{black!6}53.43 / 52.37 & \cellcolor{black!6}52.76 / 32.55 & \cellcolor{black!6}56.13 / 50.74 & \cellcolor{black!6}50.12 / 33.46 & \cellcolor{black!6}55.12 / 49.86 \\
\cmidrule(lr){2-10}
& ST3D~\cite{Yang2021ST3D} & 54.70 / 42.22 & 58.92 / 56.71 & 50.68 / 41.24 & 53.33 / 51.05 & 58.12 / 34.95 & 63.10 / 57.14 & 55.38 / 37.62 & 58.61 / 55.38 \\
& ST3D$^{\ddagger}$~\cite{Yang2021ST3D} & 54.88 / 42.85 & 59.23 / 56.40 & 51.14 / 40.88 & 55.58 / 53.27 & 56.58 / 34.58 & 63.20 / 56.02 & 53.33 / 37.57 & 56.42 / 53.29 \\
& ST3D++~\cite{Yang2023ST3DPP} & 54.84 / 42.56 & 58.15 / 56.26 & 51.06 / 41.94 & 53.56 / 52.96 & 57.13 / 34.71 & 61.93 / 56.23 & 55.56 / 39.25 & 60.96 / 55.64 \\
& ST3D++$^{\ddagger}$~\cite{Yang2023ST3DPP} & 54.82 / 43.14 & 58.62 / 56.51 & 54.78 / 43.45 & 57.48 / 55.22 & 57.16 / 34.61 & 61.94 / 56.15 & 55.51 / 39.47 & 58.70 / 55.64 \\
& ReDB~\cite{Chen2023ReDB} & 53.68 / 41.84 & 56.87 / 54.80 & 48.47 / 36.86 & 51.71 / 49.76 & 61.93 / 38.31 & 67.04 / 60.93 & 54.76 / 36.10 & 59.72 / 53.98 \\
& MS3D++~\cite{Tsai2025MS3DPP} & 55.88 / 44.07 & 60.26 / 57.54 & 50.87 / 39.91 & 55.30 / 53.08 & \textbf{64.99} / 39.69 & 69.64 / 63.57 & 60.02 / 41.20 & 63.40 / 59.98 \\
& Pi3DET-Net~\cite{Liang2025Pi3DET} & 57.22 / 47.17 & 63.69 / 61.34 & 58.76 / 47.44 & 63.68 / 61.22 & 62.64 / 44.46 & 67.69 / 63.56 & 65.82 / 49.59 & 71.22 / 65.57 \\
& \textbf{Ours} & \textbf{60.09} / \textbf{48.79} & \textbf{64.01} / \textbf{62.02} & \textbf{62.54} / \textbf{49.34} & \textbf{67.71} / \textbf{65.40} & 64.39 / \textbf{45.86} & \textbf{69.65} / \textbf{63.98} & \textbf{67.58} / \textbf{50.77} & \textbf{73.05} / \textbf{69.35} \\
\cmidrule(lr){2-10}
& Target Platform & 54.15 / 40.24 & 67.67 / 46.11 & 58.63 / 54.96 & 70.04 / 66.14 & 54.90 / 39.74 & 68.52 / 46.53 & 56.46 / 55.19 & 70.67 / 61.42 \\
\bottomrule
\end{tabular*}
}
\end{table*}

Target Simulation is applied only when Object Memory returns a valid source instance $(O_j^s,b_j^s)$ for target candidate $(O_i^t,\hat b_i)$. As detailed in Fig.~\ref{fig:simulation_detail}, the target pseudo box supplies the placement, while the retrieved source instance provides the object dimensions from its ground-truth annotation and the measured point structure from its LiDAR observation. We do not scale the source points to the dimensions of the target pseudo box.

\noindent\textbf{Viewpoint alignment.}
Let $o_s$ and $o_t$ be the source and target LiDAR origins. We use the coordinate-wise median of each object crop, denoted by $\mu_s$ and $\mu_t$, to form the sensor-to-object rays $v_s=\mu_s-o_s$ and $v_t=\mu_t-o_t$. The source crop and its ground-truth box are rotated together by the difference between the two ray azimuths:
\begin{equation}
 \Delta\psi_i=
 \operatorname{atan2}(v_{t,y},v_{t,x})-
 \operatorname{atan2}(v_{s,y},v_{s,x}).
 \label{eq:azimuth}
\end{equation}
After the yaw rotation, the box center is translated so that its bird's-eye-view center matches $(x_i,y_i)$ and its bottom face matches that of $\hat b_i$. Since the source height is retained, the resulting reference center is $\bar c_i=(x_i,y_i,z_i-h_i/2+h_j^s/2)$. For $p\in O_j^s$, the aligned point is
\begin{equation}
 \widetilde p=\bar c_i+
 R_z(\theta_j^s+\Delta\psi_i)R_x(\Delta\phi_i)
 R_z(-\theta_j^s)(p-c_j^s),
 \label{eq:point_transform}
\end{equation}
where $c_j^s$ and $\theta_j^s$ are the center and yaw of the source ground-truth box. For cross-platform pairs, $\Delta\phi_i$ is obtained from the difference between the median source and target elevation angles and clipped by a task-specific bound $\phi_{\max}$. This pitch correction is applied in the source-box coordinate system and changes only the points. The replacement box keeps the source dimensions and has horizontal heading $\theta_j^s+\Delta\psi_i$.

\noindent\textbf{Point distribution simulation and replacement.}
After viewpoint alignment, we filter the transformed source crop from the target LiDAR view. Let $\mathcal{A}_i=\{\widetilde p\mid p\in O_j^s\}$ be the aligned crop. For $q\in\mathcal{A}_i$, $\alpha(q)$ and $\varepsilon(q)$ denote the absolute azimuth and elevation deviations between the point ray $q-o_t$ and the center ray $\bar c_i-o_t$. We retain
\[
 \mathcal{V}_i=\{q\in\mathcal{A}_i\mid
 \alpha(q)\leq\tau_{\mathrm{az}},\ 
 \varepsilon(q)\leq\tau_{\mathrm{el}}\},
\]
where $\tau_{\mathrm{az}}$ and $\tau_{\mathrm{el}}$ are angular limits. If $|\mathcal{V}_i|<n_{\min}$, we retain $\mathcal{A}_i$ to avoid replacing the target observation with an overly sparse crop.

The pseudo box specifies the placement, while the retrieved source crop preserves the measured object geometry and dimensions without rescaling. Angular filtering then selects points consistent with the target view. Point coordinates remain unchanged after filtering, and the point count is not matched to that of the target crop. Cropping the retained points to the transformed source box gives $\widetilde O_i^t$. We then replace the points inside $\hat b_i$:
\begin{equation}
 \widetilde P_t^m=(P_t^m\setminus O_i^t)\cup \widetilde O_i^t.
 \label{eq:replacement}
\end{equation}
The associated pseudo label keeps the target box placement while using the source dimensions and aligned heading. Points outside the original target pseudo box are not erased, and unmatched pseudo objects are left unchanged.

\subsection{Confidence-Guided Multi-Stage Localization Reweighting}
\label{sec:cmlr}
Target Simulation corrects matched pseudo objects, but unmatched predictions are retained and their quality can vary. Cross-platform domain shifts increase false positives and localization errors; treating all pseudo objects equally would therefore give unreliable boxes the same supervision strength as reliable ones. We introduce CMLR to apply pseudo-object confidence consistently at the proposal and RoI localization stages. For every retained target pseudo object, CMLR first maps its score $s_i$ to
\begin{equation}
 \widetilde{\omega}_i=\max(s_i^\gamma,\omega_{\min}),
 \label{eq:initial_weight}
\end{equation}
\begin{equation}
 \omega_i=\beta\widetilde{\omega}_i+(1-\beta),
 \label{eq:weight}
\end{equation}
where $\gamma$ controls the sensitivity to the confidence score, $\omega_{\min}$ prevents low-score objects from losing all supervision, and $\beta\in[0,1]$ balances confidence weighting with unit weighting. Thus, high-score pseudo objects provide stronger supervision, whereas low-score objects remain in training with reduced gradient contribution. Source objects and simulated replacements use unit weight; the latter have confidence reset to $1.0$ after introducing source dimensions and measured geometry.

The same object weight is applied to proposal-stage localization and R-CNN box regression. Let $\mathcal{B}=\{\mathrm{RPN\mbox{-}loc},\mathrm{R\mbox{-}CNN\mbox{-}reg}\}$ and let $\mathcal{L}_{\mathrm{other}}$ contain the classification terms and other unchanged detector and adaptation losses. The target objective is
\begin{equation}
 \mathcal{L}_{\mathrm{ours}}=
 \mathcal{L}_{\mathrm{other}}+
 \sum_{b\in\mathcal{B}}\lambda_b
 \frac{\sum_i \omega_i\mathcal{L}_{b,i}}
 {\max(1,\sum_i \omega_i)},
 \label{eq:loss}
\end{equation}
where $\lambda_b$ is the original coefficient of branch $b$. The normalization keeps the loss scale stable across mini-batches with different numbers of pseudo objects. Source-domain ground-truth objects retain unit weight. Uncertain target predictions have less influence on optimization, and the detector architecture is unchanged.

\subsection{Training and Inference}
Adaptation starts from the source-pretrained Pi3DET-Net detector. Object Memory is built once from labeled source objects. At each scheduled pseudo-label refresh, the current detector serves as the pseudo-label teacher and predicts boxes and confidence scores for the target training scans. The gate in Eq.~\eqref{eq:retrieval_gate} pairs eligible predictions with source instances, which are then processed by Target Simulation. If no match is accepted, the original pseudo box and its points are retained. The updated pseudo labels and replacement points are reused until the next refresh.

\begin{table}[!t]
\vspace*{6pt}
\caption{Bidirectional adaptation between Pi3DET Quadruped (Q) and Drone (D). Entries are $\mathrm{AP}_{\mathrm{BEV}}/\mathrm{AP}_{\mathrm{3D}}$ (\%). Entries within each detector column share the detector and evaluation configuration; Pi3DET-Net is reproduced under this setup. ``Target Platform'' is fully supervised; the best scores among the listed adaptation methods are bolded.}
\label{tab:bidirectional_platform_results}
\centering
\scriptsize
\setlength{\tabcolsep}{0.55pt}
\renewcommand{\arraystretch}{1.03}
\begin{tabular*}{\columnwidth}{@{\extracolsep{\fill}}clcccc@{}}
\toprule
\textbf{Task} & \textbf{Method} & \multicolumn{2}{c}{\textbf{PV-RCNN}} & \multicolumn{2}{c}{\textbf{Voxel R-CNN}} \\
\cmidrule(lr){3-4}\cmidrule(lr){5-6}
& & \textbf{AP@0.7} & \textbf{AP@0.5} & \textbf{AP@0.7} & \textbf{AP@0.5} \\
\midrule
\multirow{8}{*}{\textbf{Q$\rightarrow$D}}
& Source Only & \cellcolor{black!6}27.52 / 11.76 & \cellcolor{black!6}32.60 / 27.62 & \cellcolor{black!6}28.02 / 12.14 & \cellcolor{black!6}33.93 / 28.54 \\
\cmidrule(lr){2-6}
& ST3D~\cite{Yang2021ST3D} & 31.14 / 13.93 & 38.27 / 31.90 & 30.31 / 20.06 & 36.03 / 33.05 \\
& ST3D++~\cite{Yang2023ST3DPP} & 34.35 / 16.46 & 41.97 / 36.98 & 33.88 / 20.89 & 39.29 / 36.31 \\
& ReDB~\cite{Chen2023ReDB} & 27.71 / 16.79 & 33.57 / 29.66 & 32.21 / 22.57 & 36.80 / 34.13 \\
& MS3D++~\cite{Tsai2025MS3DPP} & 31.77 / 16.47 & 40.00 / 35.27 & 26.68 / 19.43 & 29.41 / 29.20 \\
& Pi3DET-Net~\cite{Liang2025Pi3DET} & 40.62 / 26.87 & 47.60 / 44.34 & 38.04 / 20.37 & 43.74 / 40.42 \\
& \textbf{Ours} & \textbf{41.20} / \textbf{28.57} & \textbf{48.22} / \textbf{44.89} & \textbf{39.89} / \textbf{23.22} & \textbf{45.63} / \textbf{42.33} \\
\cmidrule(lr){2-6}
& Target Platform & 54.90 / 39.74 & 68.52 / 46.53 & 56.46 / 55.19 & 70.67 / 61.42 \\
\midrule
\multirow{8}{*}{\textbf{D$\rightarrow$Q}}
& Source Only & \cellcolor{black!6}39.18 / 26.46 & \cellcolor{black!6}44.90 / 40.80 & \cellcolor{black!6}35.79 / 24.85 & \cellcolor{black!6}39.99 / 37.37 \\
\cmidrule(lr){2-6}
& ST3D~\cite{Yang2021ST3D} & 50.04 / 39.16 & 54.73 / 52.36 & 46.01 / 32.18 & 51.95 / 48.14 \\
& ST3D++~\cite{Yang2023ST3DPP} & 51.80 / 40.94 & 56.29 / 53.88 & 51.15 / 35.62 & 54.30 / 51.82 \\
& ReDB~\cite{Chen2023ReDB} & 46.32 / 34.36 & 50.51 / 48.23 & 46.57 / 34.31 & 53.03 / 50.43 \\
& MS3D++~\cite{Tsai2025MS3DPP} & 50.56 / 38.46 & 55.50 / 53.13 & 49.10 / 35.09 & 52.40 / 50.28 \\
& Pi3DET-Net~\cite{Liang2025Pi3DET} & 53.14 / 41.65 & 58.87 / 56.49 & 52.96 / 38.24 & 59.98 / 56.36 \\
& \textbf{Ours} & \textbf{56.37} / \textbf{43.49} & \textbf{62.54} / \textbf{59.08} & \textbf{54.52} / \textbf{41.46} & \textbf{61.67} / \textbf{59.11} \\
\cmidrule(lr){2-6}
& Target Platform & 54.15 / 40.24 & 67.67 / 46.11 & 58.63 / 54.96 & 70.04 / 66.14 \\
\bottomrule
\end{tabular*}
\end{table}

Source and target samples are optimized with the same detector. Source supervision follows the baseline, whereas Eq.~\eqref{eq:loss} reweights the RPN localization and R-CNN regression losses of target pseudo objects; the classification losses are unchanged. At inference, only the adapted detector is used. Memory lookup, point replacement, and loss weighting are not invoked, so the inference graph is identical to that of the baseline.

\section{Experiments}
\label{sec:experiments}

\subsection{Experimental Setup}
\noindent\textbf{Datasets and evaluation.}
For the six cross-platform transfers, we follow the unsupervised Pi3DET protocol~\cite{Liang2025Pi3DET}: source annotations are available for training, whereas target annotations are used only for evaluation. The transfers span nuScenes and the Pi3DET Vehicle, Quadruped (Q), and Drone (D) platforms; we also evaluate nuScenes$\rightarrow$KITTI. All experiments use the car category and report $\mathrm{AP}_{\mathrm{BEV}}$ and $\mathrm{AP}_{\mathrm{3D}}$ over 40 recall positions. Cross-platform results use IoU thresholds of $0.7$ and $0.5$, while nuScenes$\rightarrow$KITTI follows the standard car protocol at IoU $0.7$.

\begin{table}[!t]
\vspace*{6pt}
\caption{nuScenes$\rightarrow$KITTI adaptation with PV-RCNN at IoU $0.7$. Entries are AP (\%); the best adaptation scores are bolded. ``w/ SN'' denotes statistical normalization, and ``Target Platform'' is fully supervised.}
\label{tab:cross_dataset_pvrcnn}
\centering
\small
\setlength{\tabcolsep}{2.8pt}
\renewcommand{\arraystretch}{1.09}
\begin{tabular*}{\dimexpr\columnwidth-2\tabcolsep\relax}{@{\extracolsep{\fill}}lcc@{}}
\toprule
\textbf{Method} & $\mathbf{AP}_{\mathrm{BEV}}$ & $\mathbf{AP}_{\mathrm{3D}}$ \\
\midrule
Source Only & \cellcolor{black!6}64.98 & \cellcolor{black!6}38.85 \\
\cmidrule(lr){1-3}
SN~\cite{Wang2020SN} & 55.00 & 43.95 \\
ST3D~\cite{Yang2021ST3D} & 74.47 & 50.24 \\
ST3D~\cite{Yang2021ST3D} w/ SN & 78.40 & 70.90 \\
ST3D++~\cite{Yang2023ST3DPP} & 81.70 & 45.35 \\
ST3D++~\cite{Yang2023ST3DPP} w/ SN & 84.98 & 75.50 \\
ReDB~\cite{Chen2023ReDB} & 81.26 & 50.79 \\
DTS~\cite{Hu2023DTS} & 83.90 & 71.80 \\
PERE~\cite{Zhang2024PERE} & 82.09 & 68.34 \\
CMDA~\cite{Chang2024CMDA} & 84.85 & 75.02 \\
DALI~\cite{Lu2024DALI} & 84.12 & 75.43 \\
\textbf{Ours} & \textbf{85.94} & \textbf{76.16} \\
\cmidrule(lr){1-3}
Target Platform & 88.98 & 82.50 \\
\bottomrule
\end{tabular*}
\end{table}

\begin{table}[!t]
\caption{nuScenes$\rightarrow$KITTI adaptation with Voxel R-CNN at IoU $0.7$. Entries are AP (\%); the best adaptation scores are bolded. Unreported methods are omitted, and ``Target Platform'' is fully supervised.}
\label{tab:cross_dataset_voxel}
\centering
\small
\setlength{\tabcolsep}{2.8pt}
\renewcommand{\arraystretch}{1.13}
\begin{tabular*}{\dimexpr\columnwidth-2\tabcolsep\relax}{@{\extracolsep{\fill}}lcc@{}}
\toprule
\textbf{Method} & $\mathbf{AP}_{\mathrm{BEV}}$ & $\mathbf{AP}_{\mathrm{3D}}$ \\
\midrule
Source Only & \cellcolor{black!6}41.58 & \cellcolor{black!6}17.62 \\
\cmidrule(lr){1-3}
SN~\cite{Wang2020SN} & 33.95 & 23.26 \\
ST3D~\cite{Yang2021ST3D} & 77.29 & 35.58 \\
ST3D~\cite{Yang2021ST3D} w/ SN & 87.11 & 66.02 \\
ST3D++~\cite{Yang2023ST3DPP} & 82.08 & 38.10 \\
ST3D++~\cite{Yang2023ST3DPP} w/ SN & 85.84 & 68.64 \\
ReDB~\cite{Chen2023ReDB} & 80.58 & 34.82 \\
\textbf{Ours} & \textbf{87.75} & \textbf{72.84} \\
\cmidrule(lr){1-3}
Target Platform & 91.25 & 85.29 \\
\bottomrule
\end{tabular*}
\end{table}

\noindent\textbf{Implementation.}
We implement all experiments with PyTorch~\cite{paszke2019pytorch} and OpenPCDet~\cite{openpcdet2020}, using PV-RCNN~\cite{Shi2020PVRCNN} and Voxel R-CNN~\cite{Deng2021VoxelRCNN} as in Pi3DET-Net~\cite{Liang2025Pi3DET}. Entries within each detector column share the same detector and evaluation settings; Pi3DET-Net results are our reproductions. Its official nuScenes$\rightarrow$KITTI result uses SECOND-IoU and is therefore not directly comparable to our PV-RCNN/Voxel R-CNN runs. Adaptation lasts 15 epochs with Adam~\cite{Kingma2015Adam} and a OneCycle schedule~\cite{Smith2019SuperConvergence}; pseudo-label refresh intervals follow the corresponding baseline. Object Memory retrieves $K=5$ neighbors, Target Simulation applies angular filtering, and CMLR weights RPN localization and R-CNN box regression.

\subsection{Comparison with Adaptation Methods}
Tables~\ref{tab:cross_platform_results} and~\ref{tab:bidirectional_platform_results} compare source-only training and the listed adaptation methods on the six cross-platform transfers. Table~\ref{tab:cross_platform_results} distinguishes runs with and without random object scaling (ROS), while Table~\ref{tab:bidirectional_platform_results} reports the direct Quadruped--Drone transfers. Tables~\ref{tab:cross_dataset_pvrcnn} and~\ref{tab:cross_dataset_voxel} report the nuScenes$\rightarrow$KITTI results separately for PV-RCNN and Voxel R-CNN, including the available statistical-normalization (SN) variants.

\begin{figure}[!t]
  \vspace*{5pt}
  \centering
  \includesvgasset[width=1.0\columnwidth, trim=5pt 8pt 5pt 0pt,clip]{figures/falcon_compare_00174.svg}{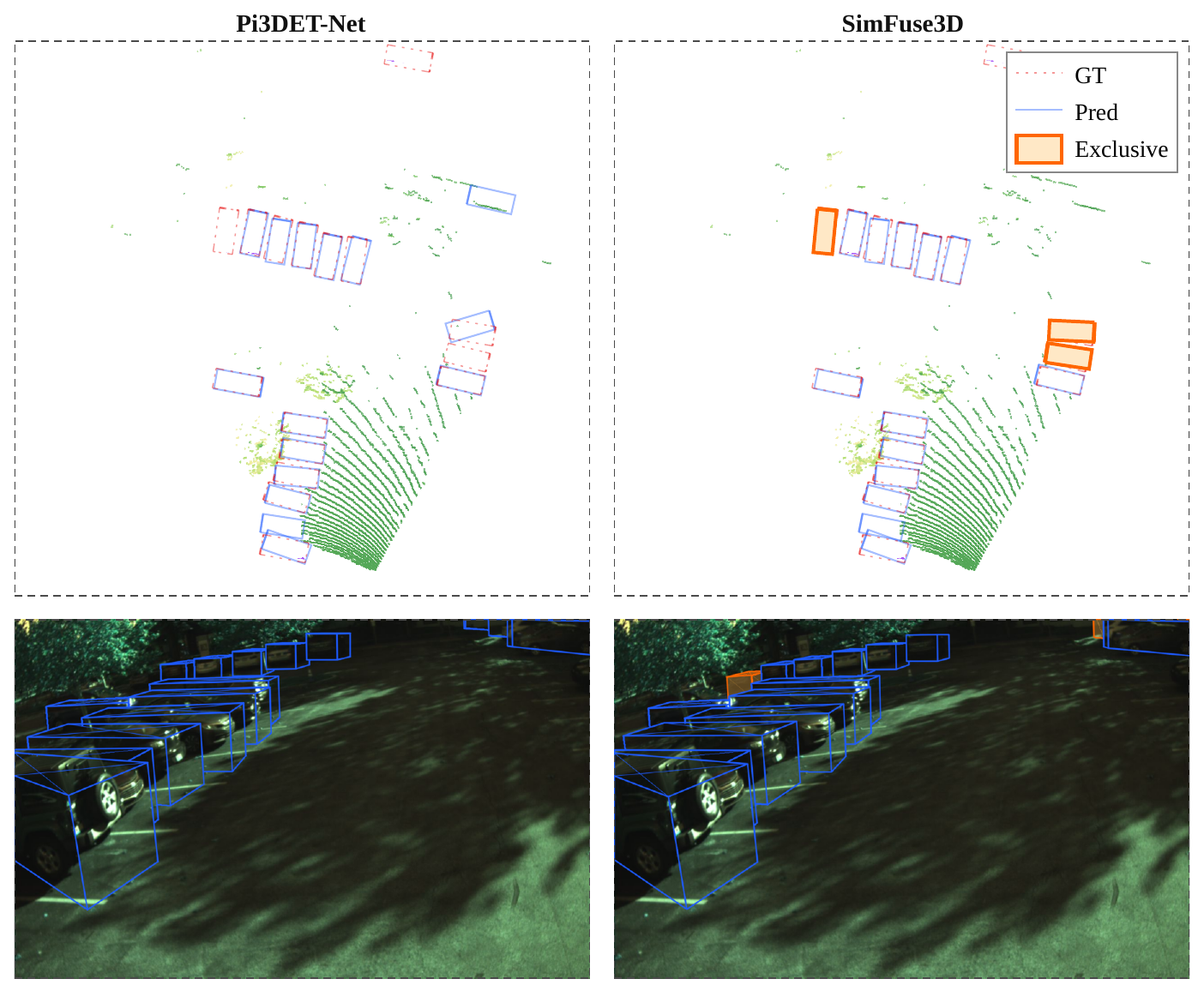}
  \caption{\textbf{Vehicle$\rightarrow$Drone comparison.} Orange outlines mark three sparse vehicles recovered by SimFuse3D at IoU $0.5$ but missed by Pi3DET-Net.}
  \label{fig:falcon_qualitative}
\end{figure}

\noindent\textbf{Cross-platform adaptation.}
Across six transfers and two detectors, SimFuse3D exceeds Pi3DET-Net in all 48 AP components and ranks first among the listed adaptation methods in 47. The exception is PV-RCNN $\mathrm{AP}_{\mathrm{BEV}}@0.7$ on Vehicle$\rightarrow$Drone, where MS3D++ obtains $64.99$ versus $64.39$. The largest margin is on nuScenes$\rightarrow$Quadruped: Voxel R-CNN $\mathrm{AP}_{\mathrm{3D}}@0.5$ rises from $55.25$ to $59.63$ ($+4.38$ points). For the two Vehicle-source transfers, all 16 components improve. Figure~\ref{fig:falcon_qualitative} shows three Vehicle$\rightarrow$Drone misses recovered over Pi3DET-Net at IoU $0.5$.

\noindent\textbf{Cross-dataset adaptation.}
On nuScenes$\rightarrow$KITTI with PV-RCNN, SimFuse3D records $85.94/76.16$, compared with $84.98/75.50$ for ST3D++ w/ SN. The gains are $0.96$ points in $\mathrm{AP}_{\mathrm{BEV}}$ and $0.66$ points in $\mathrm{AP}_{\mathrm{3D}}$. For Voxel R-CNN, the strongest adaptation-baseline values come from different rows: ST3D w/ SN reaches $87.11$ in $\mathrm{AP}_{\mathrm{BEV}}$, whereas ST3D++ w/ SN reaches $68.64$ in $\mathrm{AP}_{\mathrm{3D}}$. SimFuse3D obtains $87.75/72.84$, improving these values by $0.64$ and $4.20$ points, respectively.

\noindent\textbf{Efficiency.}
With Voxel R-CNN on Vehicle$\rightarrow$Drone, SimFuse3D uses the same detector as Pi3DET-Net and has nearly identical optimization time per epoch. Extra computation is confined to two scheduled pseudo-label refreshes, where retrieval/voting and Target Simulation add about $1.0$ min in total. Thus, 15-epoch adaptation takes $20.4$ min versus $19.2$ min for Pi3DET-Net. These operations are omitted at test time, preserving the detector and inference graph.

\begin{table}[!t]
\vspace*{6pt}
\caption{Component ablation with Voxel R-CNN on the Vehicle-source transfers. TS and CMLR denote Target Simulation and Confidence-Guided Multi-Stage Localization Reweighting. Entries are $\mathrm{AP}_{\mathrm{BEV}}/\mathrm{AP}_{\mathrm{3D}}$ (\%); best scores are bolded.}
\label{tab:ablation}
\centering
\small
\setlength{\tabcolsep}{0.8pt}
\renewcommand{\arraystretch}{1.02}
{\spaceskip=1.2pt plus 0.2pt minus 0.1pt\relax
\begin{tabular*}{\columnwidth}{@{\extracolsep{\fill}}cccccc@{}}
\toprule
\multirow{2}{*}{\textbf{TS}} & \multirow{2}{*}{\textbf{CMLR}} & \multicolumn{2}{c}{\textbf{Pi3DET (Quadruped)}} & \multicolumn{2}{c}{\textbf{Pi3DET (Drone)}} \\
\cmidrule(lr){3-4}\cmidrule(lr){5-6}
 & & \textbf{AP@0.7} & \textbf{AP@0.5} & \textbf{AP@0.7} & \textbf{AP@0.5} \\
\midrule
 &  & 58.76 / 47.44 & 63.68 / 61.22 & 65.82 / 49.59 & 71.22 / 65.57 \\
\cmidrule(lr){1-6}
$\checkmark$ &  & 60.94 / 48.22 & 65.97 / 63.58 & 67.53 / 49.66 & 70.94 / 67.36 \\
$\checkmark$ & $\checkmark$ & \textbf{62.54} / \textbf{49.34} & \textbf{67.71} / \textbf{65.40} & \textbf{67.58} / \textbf{50.77} & \textbf{73.05} / \textbf{69.35} \\
\bottomrule
\end{tabular*}
}
\end{table}

\begin{table}[!b]
\caption{Voxel R-CNN recall by target ground-truth point support. Entries are Pi3DET-Net/Ours ($\Delta$), in percent; V, Q, and D denote Vehicle, Quadruped, and Drone. \#GT is the object count, Dist. is the median sensor distance, and $\Delta$ is computed before rounding.}
\label{tab:recall_by_gt_points}
\centering
\scriptsize
\setlength{\tabcolsep}{0.65pt}
\renewcommand{\arraystretch}{1.06}
\begin{tabular*}{\columnwidth}{@{\extracolsep{\fill}}clrrcc@{}}
\toprule
\textbf{Task} & \textbf{GT pts.} & \textbf{\#GT} & \textbf{Dist. (m)} & \textbf{R@0.5} & \textbf{R@0.7} \\
\midrule
\multirow{4}{*}{\textbf{V$\rightarrow$Q}}
& 0--15     & 304   & 28.24 & 9.54/\textbf{11.51} (+1.97) & 4.93/\textbf{5.26} (+0.33) \\
& 16--30    & 180   & 25.16 & 45.56/\textbf{51.11} (+5.56) & 27.22/\textbf{30.00} (+2.78) \\
& $\geq 31$ & 1,581 & 18.46 & 80.83/\textbf{86.08} (+5.25) & 66.86/\textbf{70.46} (+3.61) \\
& All       & 2,065 & 20.41 & 67.26/\textbf{72.06} (+4.79) & 54.29/\textbf{57.34} (+3.05) \\
\midrule
\multirow{4}{*}{\textbf{V$\rightarrow$D}}
& 0--15     & 1,573 & 36.10 & 29.37/\textbf{36.62} (+7.25) & 16.59/\textbf{19.64} (+3.05) \\
& 16--30    & 871   & 28.77 & 63.15/\textbf{65.21} (+2.07) & 42.37/\textbf{44.32} (+1.95) \\
& $\geq 31$ & 3,452 & 16.01 & 85.75/\textbf{86.88} (+1.13) & 74.02/\textbf{74.10} (+0.09) \\
& All       & 5,896 & 20.36 & 67.37/\textbf{70.27} (+2.90) & 54.02/\textbf{55.17} (+1.15) \\
\bottomrule
\end{tabular*}
\end{table}

\subsection{Ablation Study}
Table~\ref{tab:ablation} reports the Voxel R-CNN ablation on the two Vehicle-source transfers. The second row adds Target Simulation (TS), and the last row further adds CMLR.

Target Simulation alone improves seven of the eight metrics, with its largest gains on Quadruped at IoU $0.5$; the only decrease is $0.28$ points in Drone $\mathrm{AP}_{\mathrm{BEV}}@0.5$. Geometry repair acts only on pseudo objects with an accepted memory match, so unmatched objects still contribute their original observations and, without CMLR, receive the same localization weight. CMLR retains these potentially correct pseudo labels but reduces the influence of lower-confidence cases. The full configuration consequently improves all eight metrics over both the TS-only row and the baseline.

\begin{figure}[!t]
  \centering
  \includesvgasset[width=1.0\columnwidth]{figures/drone_pl_s03.svg}{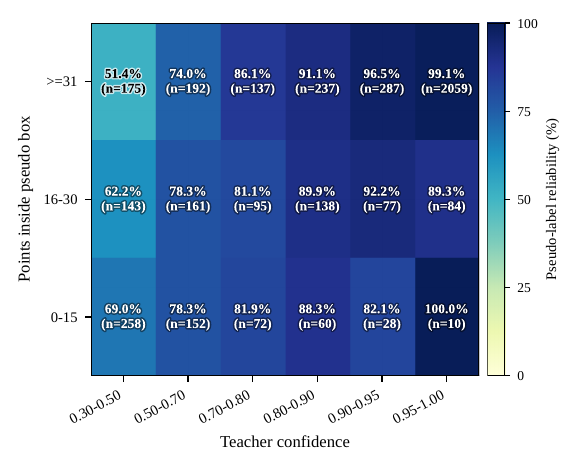}
  \caption{\textbf{Pseudo-label reliability on Vehicle$\rightarrow$Drone.} Each cell reports $Q@\mathrm{IoU}\geq0.5$ (\%) and the sample count $n$ for a teacher-confidence and in-box point-support bin.}
  \label{fig:pseudo_label_reliability}
\end{figure}

\subsection{Pseudo-label Reliability and Point-Sparse Objects}
\noindent\textbf{Reliability analysis.}
Figure~\ref{fig:pseudo_label_reliability} groups Vehicle$\rightarrow$Drone pseudo boxes by teacher confidence and in-box point count. Here, $Q$ is the fraction whose best ground-truth match reaches IoU $\geq 0.5$. Across confidence bins, $Q$ rises from $76.2\%$ for objects with 0--15 points to $93.4\%$ for those with at least 31 points, although the trend is not monotonic in every interval. Confidence and point support are therefore complementary rather than sufficient quality indicators. A failed retrieval only means that no suitable source instance was found; the original pseudo label may still provide localization supervision. SimFuse3D retains unmatched pseudo labels, downweights their localization targets with CMLR, and assigns unit weight to simulated objects.

\noindent\textbf{Recall by target point support.}
Table~\ref{tab:recall_by_gt_points} stratifies final-detector recall by target point support, with positive gains in all 16 entries. The largest gain is $7.25$ points at IoU $0.5$ for the 0--15-point Vehicle$\rightarrow$Drone bin. Overall Recall@0.5 increases by $4.79$ points on Vehicle$\rightarrow$Quadruped and $2.90$ points on Vehicle$\rightarrow$Drone. Gains at IoU $0.7$ are smaller, indicating reduced benefit under stricter overlap.

\section{Conclusion}
A useful pseudo-box location does not guarantee a reliable point observation. SimFuse3D preserves pseudo-box locations, repairs point observations with measured source geometry, and uses CMLR to adjust localization supervision.

Across six transfers, SimFuse3D improves all evaluated AP components over Pi3DET-Net by up to $4.38$ points and 
achieves the best nuScenes$\rightarrow$KITTI adaptation results among the compared methods with both detectors. Memory lookup, Target Simulation, and CMLR operate only during adaptation, leaving the deployed detector unchanged. Fixed retrieval thresholds and a stored source-object memory are limitations; adaptive retrieval and compact memory construction remain future directions.

\bibliographystyle{IEEEtran}
\bibliography{references}

\end{document}